\documentclass[apalike,iicol]{sn-jnl}

\usepackage[T1]{fontenc}
\usepackage{graphicx}
\usepackage{multicol}
\usepackage{multirow}
\usepackage{subcaption}
\usepackage{booktabs}
\usepackage{dirtytalk}
\usepackage{wrapfig}
\usepackage{comment}
\usepackage{rotating}
\usepackage{amssymb}
\usepackage{amsmath}
\usepackage{float}
\usepackage{fvextra}
\usepackage{xcolor}
\usepackage[section]{placeins}
\usepackage[square, numbers]{natbib}
\DefineVerbatimEnvironment{FancyPrompt}{Verbatim}
{
  frame=single,
  framesep=3mm,
  fontsize=\small,
  numbers=left, 
  numbersep=3pt, 
  commandchars=\\\{\}, 
  xleftmargin=10pt,
  xrightmargin=10pt,
  formatcom=\color{black}, 
  breaklines=true, 
  breakanywhere=true, 
}

\jyear{2025}
\theoremstyle{thmstyleone}%
\theoremstyle{thmstyletwo}%

\theoremstyle{thmstylethree}%

\begin{document}

\setcitestyle{square}

\title[Neurosymbolic Information Extraction]{Neurosymbolic Information Extraction from Transactional Documents\thanks{This version of the article has been accepted for publication, after peer review but is not the Version of Record and does not reflect post-acceptance improvements, or any corrections. The Version of Record is available online at: \url{https://doi.org/10.1007/s10032-025-00530-0}}}

\author[1,2]{\fnm{Arthur} \sur{Hemmer}}\email{arthur.hemmer@shift-technology.com}
\author[1]{\fnm{Mickaël} \sur{Coustaty}}\email{mcoustat@univ-lr.fr}
\author[2]{\fnm{Nicola} \sur{Bartolo}}\email{nicola.bartolo@shift-technology.com}
\author[1]{\fnm{Jean-Marc} \sur{Ogier}}\email{jmogier@univ-lr.fr}

\affil[1]{\orgdiv{L3i laboratory}, \orgname{La Rochelle University}, \orgaddress{\street{23 Av. Albert Einstein}, \city{La Rochelle}, \postcode{17000}, \country{France}}}
\affil[2]{\orgdiv{Shift Technology}, \orgaddress{\street{14 Rue Gerty Archimède}, \city{Paris}, \postcode{75012}, \country{France}}}

\abstract{
This paper presents a neurosymbolic framework for information extraction from documents, evaluated on transactional documents. We introduce a schema-based approach that integrates symbolic validation methods to enable more effective zero-shot output and knowledge distillation. The methodology uses language models to generate candidate extractions, which are then filtered through syntactic-, task-, and domain-level validation to ensure adherence to domain-specific arithmetic constraints. Our contributions include a comprehensive schema for transactional documents, relabeled datasets, and an approach for generating high-quality labels for knowledge distillation. Experimental results demonstrate significant improvements in $F_1$-scores and accuracy, highlighting the effectiveness of neurosymbolic validation in transactional document processing.

\medskip
\noindent\textbf{Publication Note:} This version of the article has been accepted for publication, after peer review but is not the Version of Record and does not reflect post-acceptance improvements, or any corrections. The Version of Record is available online at: \url{https://doi.org/10.1007/s10032-025-00530-0}
}

\maketitle

\section{Introduction}
Information Extraction (IE) is one of the foundational tasks in the field of Natural Language Processing. It involves identifying and organizing relevant pieces of information from unstructured text, which can then be used in downstream tasks, such as data mining, information retrieval, and knowledge base construction~\cite{jm3}. Like much of the history of artificial intelligence, the history of IE is a journey from labor-intensive, rule-based systems inspired by human cognition to data-driven, statistical approaches~\cite{yang2022survey}.

Early methods for information extraction heavily relied on symbolic methods such as hard-coded rules, templates, and dictionaries to extract information \cite{jones1994natural,wilks2007texttemplate,cowie-1983-automatic,dejong2014overview}. These systems were based on what were believed to be human cognitive processes. The limitations of rule-based systems quickly became apparent as they were found to be too brittle when applied to the diverse and unpredictable nature of real-world data. Maintaining these systems is labor intensive and requires continuous updates from domain experts, making them costly and inefficient. Their performance suffered significantly when faced with out-of-distribution data, demonstrating poor generalization capabilities.

Statistical methods emerged as an alternative to rule-based systems, addressing challenges in manual maintenance and rule creation. Early approaches used probabilistic models and machine learning algorithms such as Hidden Markov Models and Conditional Random Fields to learn patterns from annotated corpora for tasks like named entity recognition and relation extraction \cite{mccallum2003early,lafferty2001conditional}. With the advent of deep learning, large language models (LLMs) like GPT \cite{radford2019gpt2} have transformed information extraction. These models have improved extraction accuracy and generalization across domains, with capabilities such as few-shot learning \cite{brown2020language}. However, this progress has come with its own set of problems: LLMs lack symbolic reasoning capabilities \cite{mirzadeh2024gsm}, can confidently generate incorrect information \cite{jiang2021can,ye2023cognitive}, and require substantial computational resources and data for training and inference \cite{hoffmann2022training}. 

An ideal method for IE would have the interpretability and data-efficiency of the symbolic systems, but the generalization and flexibility of the neural methods. An emerging branch of research, known as \textit{neurosymbolics}, seeks to leverage the controlability and interpretability of rule-based systems with the scalability and adaptability of probabilistic, neural models. These methods have been most successfully applied in the field of Reinforcement Learning, as demonstrated by AlphaGo \cite{silver2016mastering}. The game's rules are symbolically enforced on the predictions of a deep neural network and the value of a move is determined in retrospect by the game's outcome, based on whether the model was victorious or not.

Games like Go and Chess are well-suited for neurosymbolic methods that rely on strict, well-defined rules, allowing for unambiguous win/loss assessment. In contrast, IE is inherently open-ended, making it challenging to establish a \say{win/loss} criteria. This complexity arises from the varied types of documents and the contextual nature of the information being extracted. In practice, however, there are many cases where there exists prior knowledge about the types of documents and even the expected content. For example, when processing documents related to a travel insurance claim, one anticipates a certain structure among the invoices, such as itemized lists of expenses, specific dates corresponding to travel events, and names of service providers like airlines or hotels. This prior knowledge can be used to serve as a validation signal, ensuring that the extracted information aligns with expected categories, formats, and relationships, such as matching expense types to relevant dates or verifying total amounts against individual charges.

In this work, we explore the integration of symbolic prior knowledge for IE from transactional documents, such as invoices and receipts. These documents adhere to a strict schema with a well-defined arithmetic relationships, enabling us to assess the coherence of extracted data. Our approach leverages the capability of schema-based validation and the open-ended, zero-shot capabilities of LLMs to produce possible extractions and filter out invalid samples. The goal of our methodology is to automatically generate high-quality labels that can be used for further downstream knowledge distillation in an offline manner, to improve model performance and efficiency.

Our key contributions include:
\begin{itemize}
    \item A comprehensive schema for transactional documents, including well-defined arithmetic relationships between fields ;
    \item Two existing datasets have been reannotated to fit this new schema schema ;
    \item A neurosymbolic methodology that combines LLMs with task- and domain-specific validation to improve information extraction accuracy.
\end{itemize}

\section{Related Work}
Information extraction from documents, and in particular transactional documents, has been a longstanding area of interest in the field of document analysis. A wide variety of benchmarks have been developed such as CORD \cite{park2019cord}, SROIE \cite{huang2019icdar2019}, WildReceipt \cite{sun2021spatial}, and DocILE \cite{vsimsa2023docile}. Although the documents in these datasets vary in type, form, and cultural context, they share a common transactional structure consisting of a list or table of sold items along with global fields in various configurations such as the total amount and the tax amount.

Despite the resemblance in terms of underlying structure, these datasets have been annotated differently, varying in label granularity and naming conventions. For instance, SROIE includes only four fields (company, address, date, and total) whereas CORD features 30 fields, categorized into subgroups like menu (items), subtotal, and total. Although CORD offers more comprehensive and precise labeling, the labels lack the necessary precision to assess the arithmetic consistency among the different fields. For example, the term \textit{total price} is ambiguous because it does not specify whether taxes are included or not.

Earlier methods for IE required significant customization of models to accommodate different schemas and their nested complexities. Consequently, existing datasets are often divided into two tasks: recognition of global key information, known as key-information localization (KILE), and line-item recognition (LIR) \cite{vsimsa2023docile}. In contrast, more recent methods based on the use of LLMs offer greater flexibility and can handle both tasks simultaneously.

Recent advancements in IE have leveraged auto-regressive LLMs to generate the desired output structures directly \cite{kim2021donut,hwang2021cost,wang2022mmlayout}. Due to the open-ended nature of these models, they can produce complex, variable-length outputs without requiring architectural modifications. Some approaches, such as Donut \cite{kim2021donut}, achieve OCR-free predictions by using the image alone as input. In contrast, other methods incorporate OCR to include the document's text directly in the model's context \cite{hwang2021cost,wang2022mmlayout}. Despite these innovations, achieving high performance still largely depends on access to expensive training datasets. 

Recently, researchers have explored the application of LLMs in zero and few-shot scenarios for document IE \cite{perot2023lmdx,luo2024layoutllm,wang2023docllm}. The $F_1$-scores in these scenarios, however, remain significantly lower than those obtained through fine-tuning. To the best of our knowledge, previous studies have not investigated the potential of domain-specific knowledge for generating high-quality labels for further distillation.

\section{Neurosymbolic Information Extraction}
\label{sec:bg}
\begin{figure*}[!htbp]
    \centering
    \vspace{4mm} 
    \includegraphics[width=\textwidth]{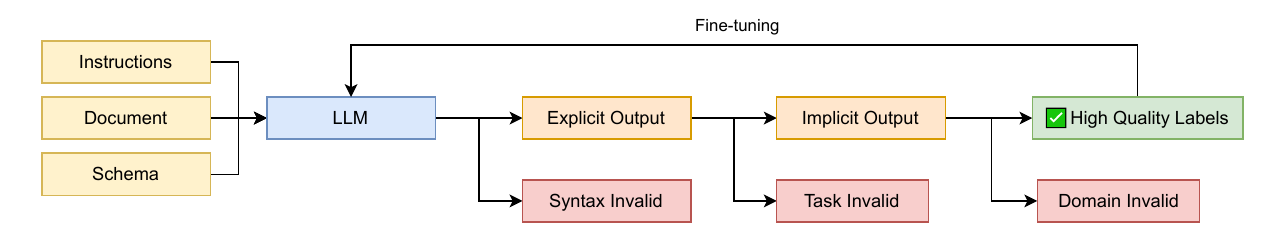}
    \vspace{4mm} 
    \caption{Overview of the information extraction pipeline, where a prompt, document, and schema are processed by an LLM. Outputs are filtered for syntax, task, and domain validity, ensuring only high-quality labels are retained}
    \label{fig:method-overview}
\end{figure*}

Fig.~\ref{fig:method-overview} provides an overview of our neurosymbolic pipeline. In this process, a LLM is given a document, in the form of the OCR-extracted text and the document image (depending on the model's capabilities). Additionally, it receives a predefined extraction schema and instructions. The model's task is to extract the requested information from the document according to the schema's format. After the model generates its predictions, multiple layers of symbolic validation are applied to filter out any erroneous outputs. This process ensures that only high-quality, accurate predictions are retained. The goal is to use these refined predictions for subsequent downstream fine-tuning, enabling effective knowledge distillation.

In the following subsections, we first formalize the general task of IE from documents. Then, we proceed to discuss the various levels of validation applied in the process.

\subsection{Structured Information Extraction}
We formalize the structured extraction task as a 3-element tuple $(\bf{x},\mathcal{S},\bf{y})$, where

\begin{itemize}
    \item $\bf{x}$ is the input document. This can be an image or the text from the document as obtained from an Optical Character Recognition (OCR) method. In the case of text, $\bf{x}$ is composed of tokens from a fixed vocabulary $\mathcal{V}$, such that $\bold{x} = \{x_1, x_2, \ldots, x_o\}$, where $x_i \in \mathcal{V}$;
    \item $\mathcal{S}$ is the expected structure, or schema, of the output, as defined further below;
    \item $\bf{y}$ is the target prediction adhering to schema $\mathcal{S}$, composed of tokens $y_i$ from the same vocabulary $\mathcal{V}$ mentionned above.
\end{itemize}

The output schema is a set of two-element tuples consisting of the unique field identifier $f_i$ (e.g., "Invoice Number", "Date", "Total Amount"), and the type of the field $\tau_i$:
\[ \mathcal{S} = \{ (f_1, \tau_1), (f_2, \tau_2), \ldots, (f_m, \tau_m) \}. \]
For nested structures, $\tau_i$ can itself be a schema $\mathcal{S}_i$, such that
\[ \tau_i = \mathcal{S}_i = \{ (f_{i1}, \tau_{i1}), (f_{i2}, \tau_{i2}), \ldots \}. \]
For lists, the type $\tau_i$ is a list of element of the same type, which again could be defined by a schema:
\[ \tau_i = \text{list of } \tau'_i. \]

An instance of the schema $\mathcal{S}$, denoted as $\mathbf{y}$, is a specific realization of the schema. It is defined as
\[ \mathbf{y} = \{ (f_1, v_1), (f_2, v_2), \ldots, (f_m, v_m) \}, \]
where we use $v_i$ to refer to the specific value extracted for field $f_i$. For information extraction, the unitary type $\tau$ is a string, but we further extend this definition for symbolic processing later on in Sec. \ref{sec:schema}.
The schema realization $\mathbf{y}$ is often serialized in a structured, machine-readable format such as JSON or XML.

\paragraph{Generative IE}
\label{sec:neural}
In generative information extraction, the task involves the sequential generation of tokens based on a comprehensive input prompt \(\mathcal{P}\). This prompt \(\mathcal{P}\) is composed of the schema \(\mathcal{S}\), specific task instructions \(\mathcal{I}\), and the document \(\mathbf{x}\), which can be given in the form of the text obtained from an OCR process and/or an image. A shortened example of a prompt can be found in App.~\ref{app:prompt}. During the generation process, each token \(t_i\) is generated conditioned on the tokens generated so far \(t_{<i}\), as well as the entire prompt \(\mathcal{P}\). This process can be expressed as:
\[
P(t_i \mid t_{<i}, \mathcal{P}) = P(t_i \mid t_{<i}, \mathcal{S}, \mathcal{I}, \mathbf{x}).
\]

The generative capability of these models allows them to produce outputs of arbitrary length and structure, handling various tasks in zero-shot or few-shot scenarios without requiring task-specific fine-tuning. However, this flexibility can introduce challenges such as generating outputs that deviate from the desired format. These challenges will be addressed in the subsequent sections.

Finally, a parsing function interprets the generated text to produce a structured output that aligns with the predefined schema \(\mathcal{S}\). This parsing process involves the use of specific delimiters or markers embedded within the generated text to map each field identifier \(f_i\) to its corresponding value \(v_i\).

\subsection{Validation}
Given that the conversion of generated tokens to structured output is strictly defined by the parse function, there may be instances where this process fails due to the model generating invalid outputs. This validation has three levels of validation employed to ensure high-quality labels:

\begin{itemize}
    \item Syntactic: This step evaluates whether the generated output follows the correct syntactic structure, checking for correctly opened and closed brackets, nested structures, and accurate field names ;
    \item Task-Level: This ensures that the extracted values are present in the original OCR text, verifying the accuracy of the extraction task ;
    \item Domain-Level: This level involves using a well-defined, arithmetically coherent structure among different fields to both complete and verify the correctness of the extraction.
\end{itemize}
Each level of validation is described in more detail below.

\paragraph{Syntactic}
The syntax of a prediction is verified by parsing the output into a JSON object. The model must generate an output that aligns with a specific JSON schema, as outlined by the JSON Schema Specification\footnote{https://json-schema.org/draft/2020-12/schema}. Even if certain fields lack corresponding values in the document, their keys must still be present in the extracted object, carrying a value of \textit{null}. For fields resembling lists, such as sub-line items with no values, an empty list is expected.

Our experiments revealed that although models typically produce JSON that conforms to the schema, they often include extraneous explanatory text or back ticks. To address this issue, we specifically parse the output enclosed between the first and last opening braces.

\paragraph{Task}
In the general task of document IE, our goal is to accurately identify and structure values from the document. As such, the second level of validation checks whether each extracted value appears verbatim in the OCR text. This constraint guards against generating information absent from the original document, preventing hallucinations.

It is important to highlight that this type of validation assumes extracted values can be found as contiguous spans within the OCR text. While this assumption generally holds true for short values, such as amounts, it can pose challenges for longer values, like line item names and addresses. Successful extraction in such cases depends on the OCR accurately capturing several bounding boxes in the correct sequence. This did not pose any problems in our experiments.

We should also consider that if an error occurs in the OCR, the model will not be permitted to rectify it, even if it successfully corrects the OCR from the image it received. In practice, this constraint does not present significant issues, as OCR error rates for transactional documents are relatively low \cite{hemmer2024confidence}. Moreover, off-the-shelf OCR systems remain more adept at reading documents than vision-language models \cite{shi2023exploring}.

\paragraph{Domain}
Domain constraints ensure that extracted information aligns with the specific rules and relationships inherent to the transactional domain. This involves maintaining arithmetic coherence among fields like totals, taxes, and line item amounts. Once the extractions are generated, we apply domain-specific symbolic processing to verify the consistency and validity of these values. This step confirms that the structured output not only captures essential information but also upholds the logical and numerical relationships expected in invoice documents. Details of the schema and evaluation process are further elaborated in the following section.

\section{Transactional Documents}
\label{sec:schema}
A transactional document is a specific type of document that represents a commercial record created by a seller and given to a buyer, listing the details of a transaction for products or services rendered. Typically, and in the context of this paper, these are invoices and receipts (for context for the reader, the technical difference is that an invoice requests payment for goods or services, while a receipt confirms payment has been received).
While they are quite similar structurally, often consisting of a list of items along with some global fields such as the total amount and the taxes, they can significantly differ in form and detail. Receipts are often small and densely packed with little detail, while invoices are generally larger and can contain a more detailed breakdown of its different components.

At the core of these documents are the numbers and amounts that make up the transaction. While these numbers can be laid out visually in many different ways, there is a strict, standardized arithmetic structure that ties them together. For example, a subtotal is the sum of the individual line items, and the gross total is the sum of the net total and the taxes. To apply domain-specific constraints to our predictions, we focused on creating a strict, unambiguous schema with well-defined arithmetic relationships between fields. 

\begin{figure*}[!htbp]
    \centering
    \vspace{4mm} 
    \includegraphics[width=\textwidth]{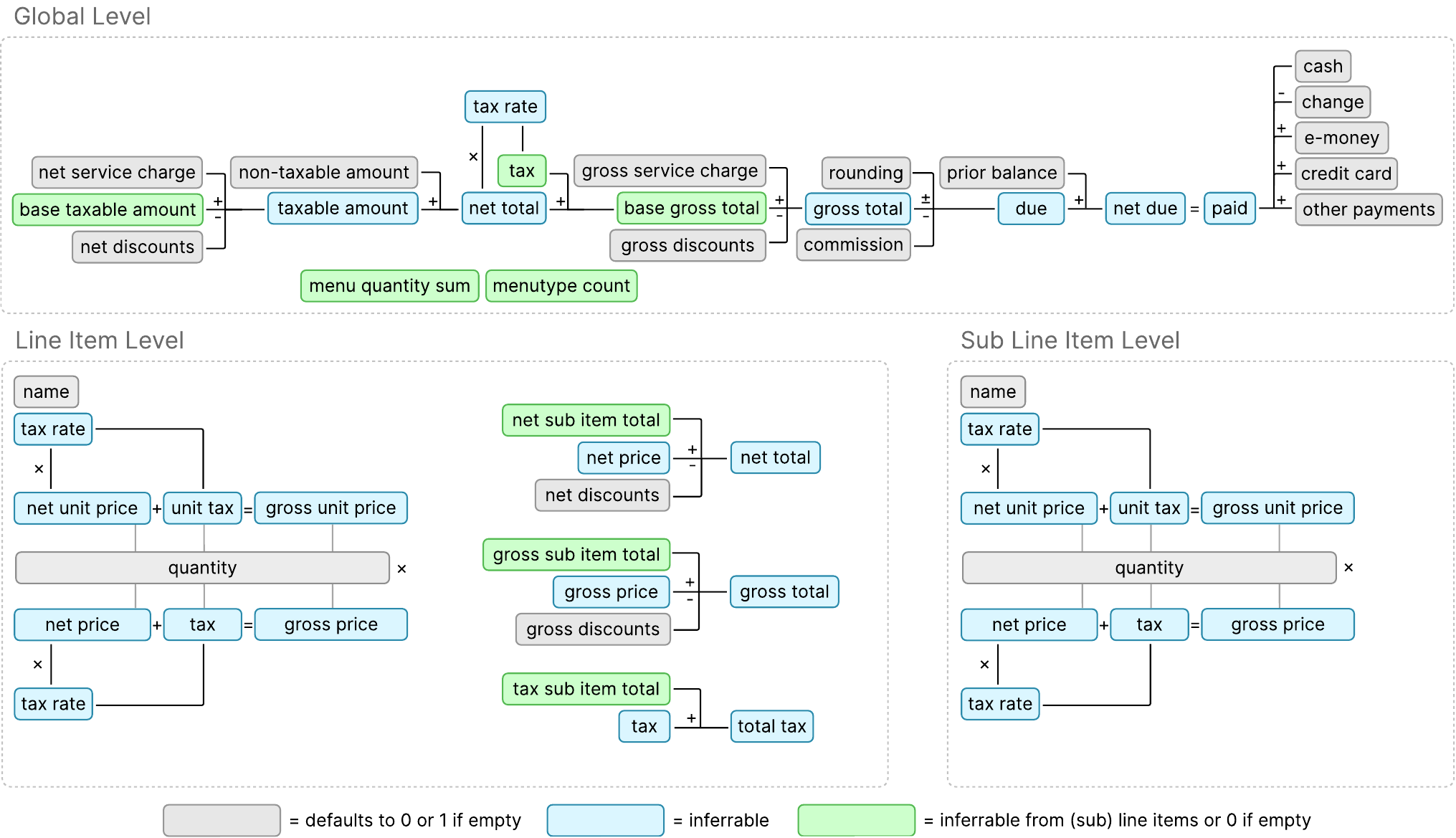}
    \vspace{4mm} 
    \caption{Overview of the Schema we introduce for Domain-level validation of the information extracted from transactional documents.}
    \label{fig:invoice-schema}
\end{figure*}

\paragraph{Schema}
The schema used is visually represented in Fig.~\ref{fig:invoice-schema}. It has a total of 53 fields and consists of three main objects, each in a one-to-many relationship to the next one in order: global, line items and sub line items. At each level there are a number of fields that are arithmetically linked. Furthermore, several fields are arithmetically linked to nested fields using summation. For example, the \textit{base taxable amount} from the global level can be inferred by summing the \textit{net total} amounts of all the line items.

\paragraph{Inference}
Except for item names (strings) and the menutype count (integers), all fields are decimal values. However, the conversion of strings to decimal values is done programmatically. The extraction predicted by the model is expected to only return string values for all fields, which we call the \textit{explicit} output. The output obtained after converting all the string values to their respective types (decimals) is called the \textit{implicit} output.

Once all values are parsed, we try to resolve any values in the schema that can be inferred. For example, when we know the net total and the tax rate, we can infer the total tax amount. We do the resolving iteratively until no more values can be inferred. When comparing predictions to the ground truth labels, we compare the resolved, implicit objects of both the prediction and the ground truth.

As shown in the schema overview, there are several fields for which we assume a default value (typically 0 or 1 in the case of quantities) if no value was provided. This helps for inference as some values are rarely present, such as commission and prior balance, and as such the gross total is often simply equal to the due and net due amounts. Having the model extract the same value for one or the other does not matter.

An important limitation of this programmatic parsing approach is the challenge of accommodating different number formatting conventions, where periods and commas are used interchangeably. This issue is compounded by variability introduced by OCR noise affecting these characters. Although we ultimately achieved an acceptable parsing rate that allows us to have >95\% of documents correctly parsed, future work could explore alternative methods for parsing these values, potentially employing another call to a language model.

\paragraph{Evaluation}
After extraction, parsing, and resolving, we can apply constraints to the implicit outputs to ensure the coherence of the extracted values. These constraints are based on the arithmetic structure of the schema and verify the equations. A document is valid if all constraints that can be evaluated are satisfied. A constraint can be evaluated if all fields in its equation have a value. If any fields lack values, the constraint is not evaluated and is omitted from the document's evaluation. 

Beyond the arithmetic constraints, several additional requirements are added to ensure the accuracy of the extraction. For example, each prediction must include at least one line item, at least a gross or net total should be extracted, and the tax rate must be between 0 and 1. Regarding the arithmetic constraints, we introduce some flexibility by checking if the two sides of the equation are approximately equal, allowing for a relative tolerance of 0.5\%. This accounts for any implicit rounding that might not be evident in the figures on the invoice.

\paragraph{Annotation}
Finally, using the developed schema, we curate several datasets by re-annotating the existing datasets CORD and SROIE. These datasets were created by mapping the existing labels to the new schema, and correcting any issues that were raised when evaluating the constraints, as well as annotating any fields that were not originally created. This correction was done by a single annotator in a few days. Both datasets are annotated with the same schema and have constraint-satisfaction rates 96.5\% (CORD) and 95.5\% (SROIE), meaning that there are 3.5\% and 4.5\% of documents respectively that do not satisfy all the constraints as expressed in the schema.

The main reasons for the non-satisfaction of constraints for these documents are (in order of most to least occurring):
\begin{itemize}
    \item Parsing issues: where interchangeably used periods and commas result in an wrong corresponding decimal value;
    \item Schema limitations: The schema may not accommodate the arithmetic structure of certain invoices. For instance, a discount might be displayed solely as a rate without the corresponding absolute value;
    \item Cut-off or readability issues: Some necessary values for coherent arithmetic may be unreadable. This can occur if they have been cropped from the image or if they've faded or the image quality is too poor;
    \item Incorrect arithmetic: A small but noticeable number of receipts feature incorrect arithmetic, such as line items that do not add up correctly or instances of excessive rounding,
\end{itemize}
The relabeled datasets are published and available for downloaded. To avoid confusion with the original datasets, we will refer to the relabeled datasets as $\text{CORD}_\textit{TD}$ and $\text{SROIE}_\textit{TD}$, where TD refers to the transactional documents schema.

\paragraph{Schema Limitations}
We have tried to make the most complete possible schema for transactional documents. For all datasets we achieve a constraint satisfaction rate of at least 95\%. To our knowledge, the most limiting factor of the schema is that it only allows for a single tax rate besides non-taxable amounts. This didn't pose any problems for the datasets we annotated, but might for other datasets. The code for the schema, inference and constraints is published along with this work.

\section{Experiments and Results}
In this section, we describe the implementation of the neurosymbolic pipeline outlined earlier. Our goal is to evaluate how different validation levels contribute to data quality. First, we examine the impact of these validations on key metrics in both zero-shot and fine-tuned settings. Then, we explore fine-tuning models on data subsets of varying quality to test whether constraints can aid in knowledge distillation.

\begin{table*}[t!]
\centering
\small
\resizebox{\textwidth}{!}{%
\begin{tabular}{lllrrrrr}
\toprule
 &  &  & \% Remaining & $F_1$ (↑) & nTED (↓) & Valid (↑) & Acc. (↑) \\
Dataset & Model & Filter &  &  &  &  &  \\
\midrule
\multirow[t]{8}{*}{$\text{CORD}_\textit{TD}$} & \multirow[t]{4}{*}{Ministral-8B} & Base & 100.0 & 60.0 & 37.8 & 28.0 & 0.0 \\
 &  &  - Syntactic & 100.0 & 60.0 & 37.8 & 28.0 & 0.0 \\
 &  &  - Task & 76.0 & 65.0 & 33.0 & 32.9 & 0.0 \\
 &  &  - Domain & 25.0 & 69.3 & 33.3 & 100.0 & 0.0 \\
\cline{2-8}
 & \multirow[t]{4}{*}{Pixtral-12B} & Base & 100.0 & 69.0 & 30.1 & 49.0 & 2.0 \\
 &  &  - Syntactic & 100.0 & 69.0 & 30.1 & 49.0 & 2.0 \\
 &  &  - Task & 83.0 & 70.3 & 29.1 & 53.0 & 2.4 \\
 &  &  - Domain & 44.0 & 74.0 & 23.8 & 100.0 & 4.5 \\
\cline{1-8} \cline{2-8}
\multirow[t]{8}{*}{$\text{SROIE}_\textit{TD}$} & \multirow[t]{4}{*}{Ministral-8B} & Base & 100.0 & 39.4 & 18.3 & 25.0 & 0.0 \\
 &  &  - Syntactic & 99.1 & 39.6 & 17.6 & 25.2 & 0.0 \\
 &  &  - Task & 41.0 & 48.1 & 15.1 & 28.7 & 0.0 \\
 &  &  - Domain & 11.7 & 52.7 & 14.7 & 100.0 & 0.0 \\
\cline{2-8}
 & \multirow[t]{4}{*}{Pixtral-12B} & Base & 100.0 & 40.5 & 15.6 & 26.8 & 0.0 \\
 &  &  - Syntactic & 100.0 & 40.5 & 15.6 & 26.8 & 0.0 \\
 &  &  - Task & 55.7 & 42.9 & 13.9 & 35.7 & 0.0 \\
 &  &  - Domain & 19.9 & 45.4 & 14.8 & 100.0 & 0.0 \\
\bottomrule
\end{tabular}
}
\caption{Zero-shot information extraction results. The Base row indicates the base model performance. Each row below filters out documents according to an increasingly strict filter. Sytactic looks at whether the output is correctly formatted, task at whether the extracted values are present verbatim in the OCR text, and domain verifies whether the parsed and inferred extraction satisfies the invoice-domain constraints}
\label{tab:firstresults}
\end{table*}

\paragraph{Metrics}
The prediction accuracy is measured with the implicit micro $F_1$-score. This score is determined by converting explicit outputs into an implicit form and resolving any remaining values, for both ground truth labels and predictions. Since the $F_1$-score does not consider the proper grouping and sequence of list items, we also use the normalized Tree Edit Distance (nTED) metric~\citep{hwang2021nted}. For calculating these metrics, we use the code provided in the Donut implementation\footnote{https://github.com/clovaai/donut/tree/1.0.7} \cite{kim2021donut}.

Additionally, we state the percentage of extractions that meet domain-specific constraints (referred to as Valid) and the full document accuracy (Doc. Acc.), which indicates the percentage of documents where all fields were extracted correctly. Lastly, for each layer of validation, we also note the percentage of documents remaining after applying the respective validation (\% Remaining). All metrics are computed strictly on this subset of remaining documents.

\paragraph{Prompt}
The instructions for the extraction task prompt are detailed in App.~\ref{app:prompt}. As outlined in the methodology section, the prompt includes the OCR text from the document, the document image (if the model supports it), the expected output schema, and additional extraction task guidelines. We did not employ advanced prompting techniques such as few-shot learning \cite{brown2020language}. Our experiments demonstrate the value of validation with zero-shot and fine-tuned models. We argue that fine-tuning is comparable to few-shot learning and anticipate similar results. In any case, we speculate that any improvements to the predictions of the model should be complementary to the symbolic validation, which we partially validate in Sec.~\ref{sec:ft}. 

The OCR text is obtained by running DocTR\footnote{https://github.com/mindee/doctr/releases/tag/v0.9.0} on each document and joining all the words together with a blank space.

\paragraph{Models}
We use Ministral 8B Instruct 2410, which is text-only, and Pixtral 12B 2409, which supports both text and images. We selected these models due to their recency and their suitability for OCR tasks, particularly the Pixtral model. Ministral 8B, being one of the newer yet smaller models, allows for fine-tuning with lower computational requirements.

All models are run at full 16-bit floating point precision using vLLM\footnote{https://github.com/vllm-project/vllm/releases/tag/v0.7.2} on a single NVIDIA H100 NVL. Unless specified otherwise, all generations are done using greedy, top-1 decoding. 

\subsection{Zero-Shot}
In this section, we examine the zero-shot setting and apply various levels of symbolic validation methods. Tab.~\ref{tab:firstresults} presents the performance of each model and dataset as filters are incrementally applied.

We observe that, with each additional layer of symbolic validation, the $F_1$-score increases across all datasets and models. This increase is relatively more pronounced for the smaller Ministral-8B compared to Pixtral-12B. However, this improvement in $F_1$ comes at the cost of filtering out a large number of documents, with Ministral retaining only 25\% and 11.7\% of documents after domain-specific filtering, compared to 44\% and 19.9\% for Pixtral. We observe that all filtering occurs at the task and domain layers, as the models are capable of generating syntactically correct outputs. The task filter removes a varying number of documents, resulting in a corresponding increase in $F_1$. However, the impact of the task filtering on document accuracy remains marginal. Retaining only documents that satisfy the domain constraints ensures an even higher $F_1$-score.

Although all remaining documents are valid after applying the domain constraints, it should be noted that the $F_1$-score does not reach 100\%. This situation arises when a model extracts only a limited number of fields, leaving much of the task to parsed inference. For instance, a model might extract only line items and not global fields. In such cases, the global fields are inferred from the line items, ensuring their validity. However, the more fields a model attempts to predict or extract, the greater the likelihood of encountering a constraint violation.

Overall, we observe lower scores for SROIE compared to CORD, which can be attributed to the greater complexity of receipts in the SROIE dataset, including more intricate tax structures. Additionally, in the CORD dataset, any content not part of the line items or global fields has been blurred out, resulting in cleaner OCR text for CORD receipts and making data extraction more straightforward. In contrast, the SROIE receipts contain more unrelated surrounding text, rendering extraction more complicated.

\subsection{Upper Bound}
\label{sec:ft}
In order to establish an upper bound and the effectiveness of our approach on more capable models, we also evaluate a fine-tuned Ministral-8B on the ground truth training labels. The results can be found in Tab.~\ref{tab:finetuned-baseline}.

\begin{table*}[t!]
    \centering
    \small
    \resizebox{\textwidth}{!}{%
        \begin{tabular}{lllrrrrr}
        \toprule
         &  &  & \% Remaining & $F_1$ (↑) & nTED (↓) & Valid (↑) & Doc. Acc. (↑) \\
        Dataset & Model & Filter &  &  &  &  &  \\
        \midrule
        \multirow[t]{4}{*}{$\text{CORD}_\textit{TD}$} & \multirow[t]{4}{*}{Ministral-8B Finetuned (GT)} & Base & 100.0 & 95.7 & 1.9 & 85.0 & 74.0 \\
         &  &  - Syntactic & 100.0 & 95.7 & 1.9 & 85.0 & 74.0 \\
         &  &  - Task & 100.0 & 95.7 & 1.9 & 85.0 & 74.0 \\
         &  &  - Domain & 85.0 & 98.1 & 1.1 & 100.0 & 83.5 \\
        \cline{1-8} \cline{2-8}
        \multirow[t]{4}{*}{$\text{SROIE}_\textit{TD}$} & \multirow[t]{4}{*}{Ministral-8B Finetuned (GT)} & Base & 100.0 & 80.6 & 2.6 & 74.7 & 16.9 \\
         &  &  - Syntactic & 100.0 & 80.6 & 2.6 & 74.7 & 16.9 \\
         &  &  - Task & 61.1 & 83.0 & 2.4 & 75.9 & 19.7 \\
         &  &  - Domain & 46.4 & 90.6 & 1.3 & 100.0 & 25.3 \\
                \bottomrule
        \end{tabular}
    }
    \caption{Results of the fine-tuned models on the ground-truth labels}
    \label{tab:finetuned-baseline}
\end{table*}

The results indicate that, similar to the zero-shot setting, the various layers of validation effectively help in selecting a subset of predictions with improved responses. The task and domain validation layers continue to prove their effectiveness by enhancing the $F_1$-score of the selected subset by approximately 10 points for SROIE and 2.4 for CORD. Additionally, these validation layers filter out significantly fewer samples compared to the zero-shot setting, showing that our method remains effective, even when considering models fine-tuned for the task at hand.

\begin{table*}[t!]
    \centering
    \small
    \resizebox{\textwidth}{!}{%
        \begin{tabular}{llllrrrrr}
        \toprule
         &  &  &  & \% Remaining & $F_1$ (↑) & nTED (↓) & Valid (↑) & Doc. Acc. (↑) \\
        Dataset & Model & Beams & Filter &  &  &  &  &  \\
        \midrule
        \multirow[t]{16}{*}{$\text{CORD}_\textit{TD}$} & \multirow[t]{16}{*}{Pixtral-12B} & \multirow[t]{4}{*}{1} & Base & 100.0 & 69.0 & 30.1 & 49.0 & 2.0 \\
         &  &  &  - Syntactic & 100.0 & 69.0 & 30.1 & 49.0 & 2.0 \\
         &  &  &  - Task & 83.0 & 70.3 & 29.1 & 53.0 & 2.4 \\
         &  &  &  - Domain & 44.0 & 74.0 & 23.8 & 100.0 & 4.5 \\
        \cline{3-9}
         &  & \multirow[t]{4}{*}{4} & Base & 100.0 & 60.4 & 43.0 & 41.0 & 2.0 \\
         &  &  &  - Syntactic & 100.0 & 62.6 & 39.7 & 44.0 & 3.0 \\
         &  &  &  - Task & 95.0 & 60.1 & 42.3 & 41.1 & 3.2 \\
         &  &  &  - Domain & 60.0 & 71.3 & 25.0 & 100.0 & 5.0 \\
        \cline{3-9}
         &  & \multirow[t]{4}{*}{8} & Base & 100.0 & 59.3 & 46.6 & 43.0 & 1.0 \\
         &  &  &  - Syntactic & 100.0 & 60.5 & 44.6 & 44.0 & 1.0 \\
         &  &  &  - Task & 98.0 & 60.6 & 44.3 & 43.9 & 1.0 \\
         &  &  &  - Domain & 76.0 & 74.5 & 22.1 & 100.0 & 3.9 \\
        \cline{3-9}
         &  & \multirow[t]{4}{*}{16} & Base & 100.0 & 57.3 & 50.5 & 39.0 & 2.0 \\
         &  &  &  - Syntactic & 100.0 & 59.8 & 44.2 & 41.0 & 2.0 \\
         &  &  &  - Task & 99.0 & 60.4 & 44.5 & 42.4 & 2.0 \\
         &  &  &  - Domain & 83.0 & 71.5 & 23.9 & 100.0 & 3.6 \\
        \cline{1-9} \cline{2-9} \cline{3-9}
        \multirow[t]{16}{*}{$\text{SROIE}_\textit{TD}$} & \multirow[t]{16}{*}{Pixtral-12B} & \multirow[t]{4}{*}{1} & Base & 100.0 & 40.5 & 15.6 & 26.8 & 0.0 \\
         &  &  &  - Syntactic & 100.0 & 40.5 & 15.6 & 26.8 & 0.0 \\
         &  &  &  - Task & 55.7 & 42.9 & 13.9 & 35.7 & 0.0 \\
         &  &  &  - Domain & 19.9 & 45.4 & 14.8 & 100.0 & 0.0 \\
        \cline{3-9}
         &  & \multirow[t]{4}{*}{4} & Base & 100.0 & 28.9 & 52.0 & 16.6 & 0.0 \\
         &  &  &  - Syntactic & 100.0 & 34.4 & 35.9 & 21.4 & 0.0 \\
         &  &  &  - Task & 83.1 & 31.8 & 50.1 & 21.4 & 0.0 \\
         &  &  &  - Domain & 27.4 & 45.5 & 18.4 & 100.0 & 0.0 \\
        \cline{3-9}
         &  & \multirow[t]{4}{*}{8} & Base & 100.0 & 24.1 & 66.9 & 11.4 & 0.0 \\
         &  &  &  - Syntactic & 100.0 & 30.5 & 51.2 & 16.3 & 0.0 \\
         &  &  &  - Task & 95.2 & 27.0 & 59.7 & 16.1 & 0.0 \\
         &  &  &  - Domain & 42.8 & 44.1 & 17.9 & 100.0 & 0.0 \\
        \cline{3-9}
         &  & \multirow[t]{4}{*}{16} & Base & 100.0 & 16.7 & 79.7 & 7.5 & 0.0 \\
         &  &  &  - Syntactic & 100.0 & 26.6 & 61.1 & 14.8 & 0.0 \\
         &  &  &  - Task & 100.0 & 22.8 & 68.1 & 13.0 & 0.0 \\
         &  &  &  - Domain & 51.2 & 41.9 & 18.1 & 100.0 & 0.0 \\
        \bottomrule
        \end{tabular}
    }
    \caption{Impact of sampling on the effectiveness of the different levels of validation and filtering. Samples were drawn with a sampling temperature $\tau=1$}
    \label{tab:sampling}
\end{table*}

\subsection{Sampling}
While generally for structured information extraction one would want to avoid variance, having a symbolic validation method allows us to use this as a strength by sampling multiple responses and selecting the ones that are valid. We investigate the usefulness of sampling multiple responses.

We sample a number of predictions per document at a temperature of 1.0. If, after validation, there are multiple valid predictions for a single sample, we take the prediction with the highest average probability over all predicted tokens. The results are shown in Tab.~\ref{tab:sampling}. All the values shown in the table are calculated per document, not per sample. This means that if all the samples for a document are invalid, it only counts as one invalid result.

Overall, we observe that increasing the sample size generally leads to a decrease in the base $F_1$-score, although this trend fluctuates. Specifically, for CORD, there's a decrease of 11.7 points between 1 and 16 samples, and for SROIE, a drop of 23.8 points. The variability in the $F_1$-score is more pronounced in domain-specific scores. For instance, in the CORD dataset, the score drops to 71.3 at 4 samples before rising to 74.5 at 8 samples. This variability can be attributed to randomness in sampling, which we found challenging to quantify through repeated experiments due to computational limitations. Consequently, when sampling, we do not anticipate a significant increase in the $F_1$-score at any of the validation levels.

However, in terms of recall, we observe a positive trend in both the percentage of documents retained. In the CORD dataset, while the domain $F_1$-scores remains relatively steady, the percentage of retained documents almost doubles when comparing greedy sampling to sampling with 16 samples. For SROIE, the percentage of remaining documents after the domain filter increases by 157\%. This indicates that despite the challenges associated with sampling, it can be consistently used to improve recall.

\subsection{Knowledge Distillation}
Earlier, we implemented various levels of constraints to isolate a subset of high-quality extractions. In the final phase of our experiments, we assess the effectiveness of these high-quality labels for knowledge distillation. Using the predictions from Pixtral-12B, we fine-tune various models on both the unfiltered and filtered predictions to measure the impact. For this experiment, we fine-tune the aforementioned Ministral-8B, Llama-3.1-8B-Instruct~\cite{dubey2024llama3}, and Qwen2.5-14B-Instruct~\cite{qwen2,qwen2.5}.

Since the base models have not previously been trained on the new datasets, we use the training set to generate the labels for fine-tuning. The models are fine-tuned using LoRA~\cite{hu2022lora}, employing a rank and alpha of 128, a dropout rate of 5\%, a learning rate of 5e-5 with a warmup ratio of 10\%, and a cosine schedule. Each model is trained for 4 epochs, with 4 evaluations per epoch on the validation set, which is a random 20\% split from the training set. The checkpoint with the lowest validation loss is retained as the final model. The results are shown in Tab.~\ref{tab:distillation}.

Overall, we observe that the models fine-tuned on the domain-filtered subsets achieve better extraction results than their counterparts fine-tuned on the unfiltered, base datasets. Notably, on CORD, Ministral-8B improves its $F_1$-score from 69.7 to 77.4 by fine-tuning solely on the filtered predictions, which is higher than the $F_1$-score measured on the filtered predictions (74.0). However, this seems to be an outlier, as the other models show less significant increases in $F_1$-score.

On SROIE, we also observe consistent improvements in $F_1$-score when fine-tuning on the domain-specific subset, with improvements ranging from 2.1 to 2.5. Similar to Ministral-8B on CORD, we see that Qwen-2.5-14B achieves an $F_1$-score higher than the $F_1$-score measured on the domain subset used for training. However, the increased complexity of SROIE results in relatively low overall zero-shot scores compared to CORD, and we do not manage to extract a completely accurate document.

\begin{table*}[bth]
    \centering
    \small
    \resizebox{\textwidth}{!}{%
    \begin{tabular}{lllrrrr}
    \toprule
     &  &  & $F_1$ (↑) & nTED (↓) & Valid (↑) & Doc. Acc. (↑) \\
    Dataset & Model & Subset &  &  &  &  \\
    \midrule
    \multirow[t]{6}{*}{$\text{CORD}_\textit{TD}$} & \multirow[t]{2}{*}{Ministral-8B} & Base & 69.7 & 28.0 & 62.0 & 5.0 \\
     &  & Domain & 77.4 & 23.4 & 81.0 & 6.0 \\
     \cline{2-7}
     & \multirow[t]{2}{*}{Llama-3.1-8B} & Base & 69.2 & 24.5 & 62.0 & 9.0 \\
     &  & Domain & 72.6 & 23.8 & 77.0 & 7.0 \\
    \cline{2-7}
     & \multirow[t]{2}{*}{Qwen-2.5-14B} & Base & 70.4 & 27.3 & 58.0 & 6.0 \\
     &  & Domain & 72.4 & 22.9 & 75.0 & 8.0 \\
    \cline{1-7} \cline{2-7}
    \multirow[t]{6}{*}{$\text{SROIE}_\textit{TD}$} & \multirow[t]{2}{*}{Ministral-8B} & All & 41.2 & 15.2 & 29.8 & 0.0 \\
     &  & Domain & 43.7 & 13.7 & 36.7 & 0.0 \\
    \cline{2-7}
    & \multirow[t]{2}{*}{Llama-3.1-8B} & Base & 42.1 & 14.8 & 32.5 & 0.0 \\
     &  & Domain & 44.2 & 14.0 & 35.8 & 0.0 \\
    \cline{2-7} 
     & \multirow[t]{2}{*}{Qwen-2.5-14B} & Base & 43.9 & 15.1 & 31.9 & 0.0 \\
     &  & Domain & 46.2 & 13.6 & 43.7 & 0.0 \\
            \bottomrule
        \end{tabular}        
    }
    \caption{Results of fine-tuning on filtered and unfiltered predictions. (Base) indicates the model fine-tuned on the Pixtral-12B unfiltered predictions and (Domain) indicates the model fine-tuned on the Pixtral-12B domain filtered predictions}
    \label{tab:distillation}
\end{table*}

\section{Conclusion}
This work demonstrates the effectiveness of a neurosymbolic approach for extracting information from transactional documents, combining LLMs with symbolic validation layers. By enforcing schema-driven constraints, we enhance extraction accuracy and achieve high-quality outputs that adhere to domain-specific rules, including arithmetic coherence. Our framework's multi-layer validation, covering syntax, task fidelity, and domain consistency, substantially improves the quality of extracted information, particularly in zero-shot and fine-tuning scenarios. The relabeled datasets and schema we provide can serve as benchmarks for further research.

\section{Limitations \& Future Work}
In addition to the limitations mentioned throughout the paper, we recognize that our methodology relies heavily on domain-specific schemas and constraints, which can limit its initial adaptability to other document types. Because of this, in our experiments, we have decoupled the different levels of constraints in a way that allows them to be separated from the strict schema. For example, the syntactic filter applies generally to all types of structured extraction, and the task filter, where we check whether the extracted values are present verbatim in the OCR output, applies to all visual IE tasks in general, without needing the strict schema. Although the syntactic constraints do not show much of a difference, the task constraints demonstrate a modest but consistent improvement over the unfiltered generations. Future work could focus on creating more domain-invariant constraints that can be used for validation.

However, as we have shown, the domain-specific constraints do add significant value to our methodology. As a further line of inquiry, future work could examine how to generate such domain-specific schemas and constraints automatically.

Finally, OCR errors have not been addressed in our current work. While these errors would not significantly affect our findings, they could be managed in future research using OCR error correction techniques~\cite{hemmer2024confidence}.

\section*{Declarations}

\bmhead{Funding}
This work has been partially funded in the framework of the France Relance. This project was also provided with computing HPC/AI and storage resources by IDRIS at CINES thanks to the grant 2023-AD010614769 on the supercomputer Jean Zay's A100/ H100 partition.

\bmhead{Competing Interests}
The authors have no financial or competing interests to declare that are relevant to the content of this article.



\footnotesize

\bibliography{manuscript}




\normalsize
\appendix

\onecolumn
\section{Prompt}

\label{app:prompt}
The prompt containing the instruction for the extraction task can be found below. 

\vspace{1em}
\begin{FancyPrompt}
  Your task is to extract the information for the fields provided below. Extract the information in JSON format according to the following JSON schema:
  \{
    "\$defs": \{
      "Invoice": \{
        "properties": \{
          "base_taxable_amount": \{
            "anyOf": [\{"type": "string"\}, \{"type": "null"\}],
            "default": null,
            "description": "The base amount that is subject to tax",
            "title": "Base Taxable Amount"
          \},
          "net_discounts": \{
            "anyOf": [\{"items": \{"type": "string"\}, "type": "array"\}, \{"type": "null"\}]
          \}
          ...
        \}
      \}
    \}
  \}
  
  Additional guidelines:
  - Extract only the elements that are present verbatim in the document text. Do NOT infer any information.
  - Extract each element EXACTLY as it appears in the document.
  - Each value in the OCR can only be used AT MOST once. If a value can correspond to multiple fields, pick the best one.
  - For each object, output all the keys from the schema even if the value is null. Empty lists should be outputted as lists with no elements.
  - If no indication of tax is given, assume the amounts to be gross amounts.
  <ocr>
    ...
    TOTAL: RP. 18,000.00
    ...
  </ocr>
  Please read the text carefully and follow the instructions.
  [If the model takes images as input]
  An image of the document is provided below.
\end{FancyPrompt}
\vspace{1em}

\end{document}